\documentclass[lettersize,journal]{IEEEtran}
\usepackage{amsmath,amsfonts}
\usepackage{algorithmic}
\usepackage{algorithm}
\usepackage{array}
\usepackage[caption=false,font=normalsize,labelfont=sf,textfont=sf]{subfig}
\usepackage{textcomp}
\usepackage{stfloats}
\usepackage{url}
\usepackage{verbatim}
\usepackage{graphicx}
\usepackage{cite}

\usepackage{xcolor}
\usepackage{tcolorbox}
\usepackage{multirow}
\usepackage{graphicx}
\usepackage{booktabs}
\usepackage{color, colortbl}
\usepackage{hyperref}
\usepackage{threeparttable}

\definecolor{Gray}{gray}{0.9}
\definecolor{demphcolor}{RGB}{144,144,144}
\definecolor{backblue}{RGB}{221,239,251}
\definecolor{backblue1}{RGB}{186,216,242}
\definecolor{backred}{RGB}{244,199,204}

\newcommand{\demph}[1]{\textcolor{demphcolor}{#1}}

\begin{document}

\title{TPCap: Unlocking Zero-Shot Image Captioning with \underline{T}rigger-Augmented and Multi-Modal \underline{P}urification Modules}

\author{Ruoyu Zhang, Lulu Wang, Yi He, Tongling Pan, Zhengtao Yu, Yingna Li~\IEEEmembership{}
	\thanks{This work was supported by the Yunnan Science and Technology Major Project(No.202302AD080002, No.202402AD080003), Yunnan Fundamental Research Projects (No.202201AS070029),  Yunnan High level Science and Technology Talents and lnnovation Team Selection Special	Project (No.202405AS350001), Yunnan Fundamental Research Projects (No.202401AU070162). (Corresponding author:Yingna Li)
		
		Ruoyu Zhang, Lulu Wang, Tongling Pan, Zhentao Yu and Yingna Li are Faculty of Information Engineering and Automation, Kunming University of Science and Technology, Kunming 650504, China
		
		Lulu Wang, Zhentao Yu and Yingna Li are Yunnan Key Laboratory of Computer Technologies Application, Kunming 650504, China
		
		Yi He is Hongyun Honghe Group Honghe Cigarette Factory, Kunming 650202, China}
	\thanks{}}

\markboth{Journal of \LaTeX\ Class Files,~Vol.~14, No.~8, August~2021}%
{Shell \MakeLowercase{\textit{et al.}}: A Sample Article Using IEEEtran.cls for IEEE Journals}


\maketitle

\begin{abstract}
Recent advancements in large language models (LLMs) have significantly enhanced the fluency and logical coherence of image captioning. Retrieval-Augmented Generation (RAG) is widely adopted to incorporate external knowledge into LLMs; however, existing RAG-based methods rely on separate retrieval banks, introducing computational overhead and limiting the utilization of LLMs' inherent zero-shot capabilities.
To address these limitations, we propose TPCap, a novel trigger-augmented and multi-modal purification framework for zero-shot image captioning without external retrieval libraries. TPCap consists of two key components: trigger-augmented (TA) generation and multi-modal purification (MP). The TA module employs a trigger projector with frozen and learnable projections to activate LLMs' contextual reasoning, enhance visual-textual alignment, and mitigate data bias. The MP module further refines the generated entity-related information by filtering noise and enhancing feature quality, ensuring more precise and factually consistent captions.
We evaluate TPCap on COCO, NoCaps, Flickr30k, and WHOOPS datasets. With only 0.82M trainable parameters and training on a single NVIDIA RTX 4090 GPU, TPCap achieves competitive performance comparable to state-of-the-art models. 
\end{abstract}

\begin{IEEEkeywords}
Image captioning, retrieval-augmented generate, large language model, zero-shot.
\end{IEEEkeywords}

\section{Introduction}
\begin{figure}[!t]
	\centering
	\includegraphics[width=0.5\textwidth]{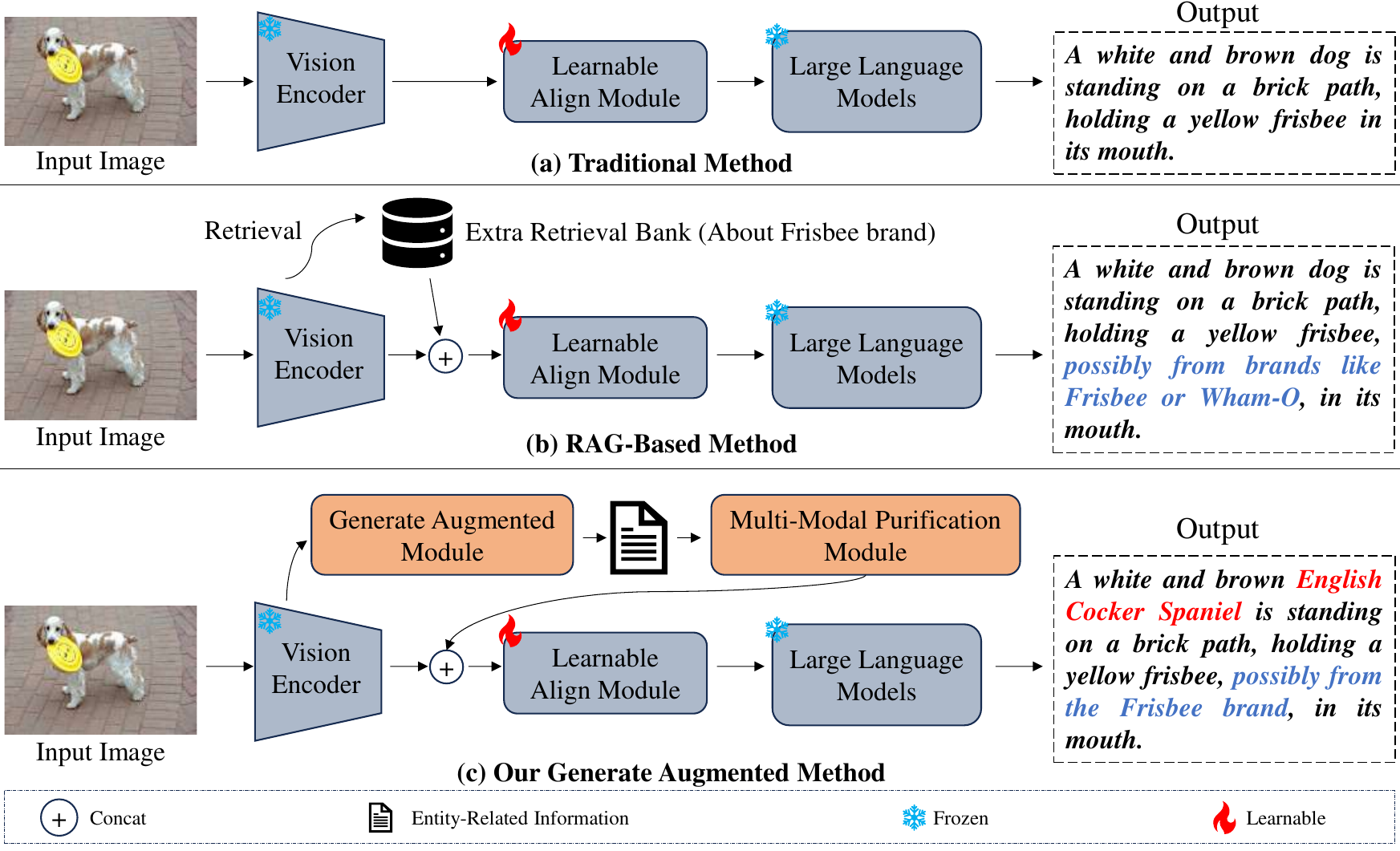}
	\caption{Comparison of different methods of using LLMs generates image captioning. \textbf{(a) Traditional methods.} Traditional methods lack detail, making it difficult to generate accurate descriptions. \textbf{(b) RAG-based methods.} This method is limited by the knowledge in the additional retrieval bank. When the additional retrieval bank is about the Frisbee brand, the generated captions can describe the Frisbee in detail, but the description of the dog is still not detailed enough. \textbf{(c) Our generate augmented method.} Our method replaces the additional retrieval bank with LLMs and generates additional information by activating the zero-shot ability of LLMs to assist in generating more detailed descriptions.}
	\label{fig_0}
\end{figure}

\IEEEPARstart{I}{mage} captioning~\cite{wang2022text, song2024embedded, zhao2023boosting} is a fundamental task in computer vision that focuses on generating natural language descriptions for images. With the rapid advancement of large language models (LLMs), researchers have increasingly integrated them into image captioning frameworks to enhance descriptive quality. This integration has led to significant improvements in cross-modal understanding, particularly in refining the logical coherence and fluency of generated captions, thereby advancing the overall performance of multimodal tasks.

Classical approaches~\cite{alayrac2022flamingo, li2023blip, liu2024visual} typically employ trainable alignment modules to bridge visual and language features. However, these methods often fail to generate accurate and detailed descriptions, as shown in Fig.~\ref{fig_0}(a). They are also prone to language bias inherent in LLMs, ignoring crucial visual cues and generating hallucinated content. To address these limitations, retrieval-augmented image captioning~\cite{hu2023reveal, shen2024imagpose, ramos2023smallcap, li2024evcap} has been proposed, combining retrieval-augmented generation (RAG)~\cite{lewis2020retrieval} with LLM-based image captioning. These methods utilize external retrieval banks to enhance descriptive accuracy by retrieving supplementary information based on visual features. However, as shown in Fig.~\ref{fig_0}(b), the reliance on pre-defined retrieval banks imposes inherent limitations, as information outside the retrieval bank cannot be captured. Furthermore, noisy retrievals from low-quality retrieval banks can mislead LLMs, degrading overall performance~\cite{shi2023large, fang2024enhancing, xu2024unsupervised}. In summary, we argue that the effective incorporation of additional information, along with improved alignment of visual and textual features, is crucial for advancing image captioning performance.

Recent works~\cite{ramos2023smallcap, yasunaga2022retrieval, shen2024imagdressing, yang2023re} highlight the increasing prominence of retrieval-augmented image captioning due to its low update cost and superior zero-shot capabilities. These approaches construct retrieval banks consisting of image-text pairs to provide auxiliary information. During inference, image features are used as queries to retrieve similar features from the retrieval bank, and their corresponding text descriptions are used to enrich captions. However, such methods heavily depend on the quality of the retrieval bank, and mismatched feature embeddings between the query and the retrieval bank can introduce semantic noise, further limiting performance.

Aligning visual and textual features has long been a critical challenge in image captioning. Traditional transformer-based methods~\cite{shen2023triplet, hu2024exploring, shen2023pbsl, Shao2022} leverage cross-modal attention to achieve alignment. However, in the LLM era, simpler strategies~\cite{li2023blip, liu2024visual} rely on learnable modules to align image and text features directly. While effective, such methods often require large-scale datasets for fine-tuning. Current LLM-based image captioning models seldom undergo extensive fine-tuning, leading to suboptimal alignment performance outside fine-tuned domains.

To address the challenges in image captioning, we introduce a novel retrieval-augmented image captioning framework (TPCap) that eliminates the need for pre-built retrieval banks. Instead, we leverage Large Language Models (LLMs) to dynamically generate retrieval-like information during inference, offering a more adaptive and scalable solution. To maintain model simplicity and facilitate efficient iteration, we minimize the number of trainable parameters, keeping the model largely frozen wherever possible.
A key innovation of our framework is the introduction of a trigger projector, consisting of a frozen projector and a trainable projector. This design balances dataset and language biases inherent in LLMs, effectively aligning visual and textual information and mitigating hallucination effects. The trigger projector ensures that TPCap has only 0.82M trainable parameters, maintaining the model's simplicity without sacrificing performance. Additionally, we develop a multi-modal purification (MP) module that refines the LLM-generated information by filtering out irrelevant noise, significantly improving the factual consistency and relevance of captions.
Extensive experiments on benchmark datasets, including COCO, NoCaps, Flickr30k, and WHOOPS, demonstrate that TPCap not only enhances captioning accuracy but also generates more detailed and contextually relevant descriptions.

The contributions of this paper can be summarized as follows:
\begin{itemize}
    \item We propose a novel retrieval-augmented image captioning framework that leverages LLMs to dynamically generate entity-related information without requiring an external retrieval bank. TPCap incorporates a \textit{multi-modal purification} (MP) module to refine generated information, eliminating the need for labor-intensive retrieval bank construction.
    
    \item We introduce a \textit{trigger projector}, combining a frozen projector and a trainable projector to balance dataset bias and language bias in LLMs. This mechanism aligns visual and textual information effectively, mitigating hallucination effects and producing captions that align with the image content. \textcolor{black}{At the same time, \textit{trigger projector} ensures that TPCap has only 0.82M trainable parameters, maintaining the simplicity of the model.}

    \item Extensive experiments on benchmark datasets, including COCO, NoCaps, Flickr30k, and WHOOPS, demonstrate that the proposed TPCap framework significantly improves captioning accuracy and generates more detailed, precise descriptions. 
\end{itemize}

\section{Related Work}
In this section, we review the literature on image captioning, retrieval-augmented generation (RAG), and visual-language alignment, highlighting their advancements and limitations.

\subsection{Image Captioning} 
Image captioning aims to generate natural language descriptions for visual content. Traditional methods adopt an encoder-decoder architecture~\cite{vinyals2015show, wu2019recall, xu2019multi} trained in an end-to-end fashion, where pre-trained visual models extract image features that are subsequently decoded into textual descriptions. Early research primarily focused on aligning visual features at different granularities with textual semantics. For instance, $M^{2}$~\cite{cornia2020meshed} employs a memory-enhanced encoding layer and a reticular connection decoding layer to model the multi-level relationships between image regions and text. DLCT~\cite{luo2021dual} integrates grid and region features to better align semantics at varying granularities with textual descriptions. HAAV~\cite{kuo2023haav} introduces heterogeneous views of input images to enrich semantic granularity, employing a shared encoder for all views to enhance alignment. VST~\cite{zhang2023improving} incorporates a global position-sensitive co-attention encoder to improve spatially-aware semantic interactions between visual and textual features.
Despite their effectiveness in enhancing visual-semantic granularity, these methods largely neglect the semantic limitations of textual features, which hinders their ability to generalize to open-world scenarios. 

\subsection{Retrieval-Augmented Generation}
Retrieval-augmented generation (RAG) enhances generation tasks by incorporating external knowledge retrieved based on the input query. Initially introduced in natural language processing (NLP)~\cite{lewis2020retrieval}, RAG improves prediction quality by retrieving relevant information from large-scale document repositories to generate more informed outputs~\cite{singh2021nlp, xu2022rag}. Recently, RAG has been extended to image captioning. For instance, AoANet~\cite{fei2021memory} utilizes a memory bank of image-sentence pairs and target words to enhance caption generation. SmallCap~\cite{ramos2023smallcap} retrieves captions from a dedicated datastore using image-to-text similarity. RA-CM3~\cite{yasunaga2022retrieval} integrates text and image information in a dense multimodal retriever to fetch relevant documents. EXTRA~\cite{ramos2023retrieval} and Re-ViLM~\cite{yang2023re} retrieve captions by comparing input image features with vision candidates in the retrieval bank.
However, these approaches require constructing and maintaining an external retrieval bank, which not only increases computational and storage overhead but also presents usability challenges for non-expert users. 

\subsection{Visual-Language Alignment}
Effective alignment of visual and textual features is critical for image captioning, as it directly influences the quality of the generated descriptions. CLIP~\cite{radford2021learning} pioneered the use of contrastive learning to align image and text features into a shared embedding space. Building on this, recent methods have explored visual-language alignment within the context of LLMs. For example, GIT~\cite{wang2022git} utilizes a generative image-to-text converter to establish alignment. BLIP-2~\cite{li2023blip} connects frozen vision encoders and LLMs through generative pre-training. Flamingo~\cite{alayrac2022flamingo} introduces GATED XATTN-DENSE layers into frozen language model layers to facilitate alignment. LLava~\cite{liu2024visual} employs a learnable projection matrix to map visual features into the same embedding space as the word embeddings of the LLM, enabling efficient alignment.
While these methods have achieved significant success, their performance often deteriorates when fine-tuned on small, biased datasets, leading to suboptimal feature alignment and biased mappings. 


\begin{figure*}[!t]
	\centering
	\includegraphics[width=1\textwidth]{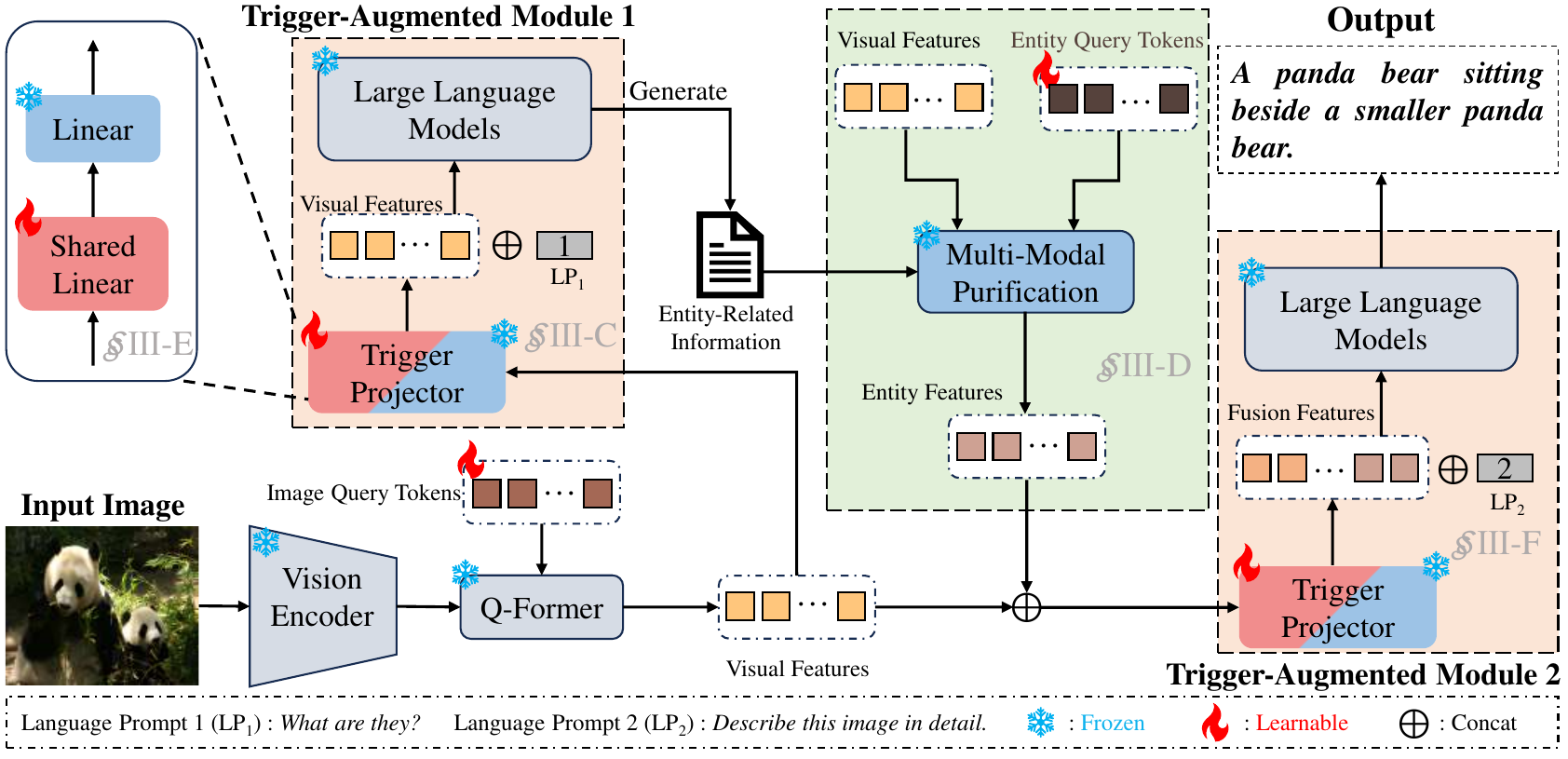}
	\caption{\textbf{Overview of the proposed TPCap.} We introduce a specialized RAG approach and a trigger projector to assist the network in aligning visual features with text features and enhancing its zero-shot capability. \textbf{First}, given an image, we extract visual features using a frozen visual encoder and generate visual-language features through a frozen Q-Former. \textbf{Then}, the visual-language features are concatenated with language prompt 1 and projected into the shared dimension by a trigger projector to enhance alignment ability. \textbf{Then}, the projected features are input into frozen LLMs to generate coarse-grained information about the entity. \textbf{Then}, a multi-modal purification is used to purify and refine the coarse-grained entity information and align it with visual-language features. \textbf{Then}, we concatenate visual-language features, entity features, and language prompt 2, projected into the shared dimension by the trigger projector to enhance alignment ability. \textbf{Finally}, a frozen LLM uses the projected features to generate output.}
	\label{fig_1}
\end{figure*}

\section{\textcolor{black}{Method}}
\subsection{Overview}
The goal of our proposed TPCap is to establish a new paradigm for retrieval-augmented generation without relying on additional retrieval banks. The TPCap architecture is shown in Fig.~\ref{fig_1}. We describe the model in six parts: image encoding, trigger-augmented module 1, multi-modal purification module, trigger projector, trigger-augmented module 2, and trainning and inference. 
Specifically, the image encoding section briefly describes our approach to extracting image features, while the entity-related trigger-augmented module 1 section explains how we generate entity-related information using trigger-augmented module 1. In the multi-modal purification module section, we mainly describe how to use the multi-modal purification module to refine the entity-related information. In the trigger projector section, we explain in detail how to use the trigger projector for visual-language alignment. In the trigger-augmented module 2 section, we elaborate on how to generate image captions using trigger-augmented module 2. Finally, we describe how to use the model for training and inference.

\subsection{Image Encoding.}
First, we extract image features to convert the image into a tensor that the model can process. Given a batch of images, a frozen pre-trained vision encoder transforms them into visual features. Then, to improve model efficiency and remove excess noise from the image features, we use the frozen pre-trained Q-Former to compress and purify the image features. At the same time, to ensure that the compression process is controllable, we set the image query tokens in Q-Former to be trainable. Finally, learned visual features are generated. The entire process is expressed as follows:
\begin{equation}
	\begin{aligned}
			& F_{img} = \varepsilon(X), \\
			& F_{v} = Q(F_{img}, T_{img}),
		\end{aligned}
\end{equation}
where $F_{img}$ represents image features, $\varepsilon$ is vision encoder, $X$ denotes images, and $F_{v}$ represents visual features, which we denote as $F_{v}=\{f_{v1},f_{v2},\cdots,f_{vn}\}$(1 $\times$ 768 each), $Q$ is the frozen pre-trained Q-Former, $T_{img}$ represents the trainable image query tokens.

\subsection{Trigger-Augmented Module 1.}
To acquire external knowledge, we first consider using traditional RAG, but RAG requires building additional retrieval libraries, which is a significant overhead and often needs to be constantly updated for different tasks.  At the same time, LLMs can be viewed as a retrieval bank, which contrasts with the traditional idea. LLMs contain a vast amount of knowledge but require specific methods to guide the generation.  Therefore, we propose a special RAG that generates entity features by leveraging the zero-shot capability of LLMs. To achieve this, we propose a trigger-augmented module 1 (TA1) to generate entity-related information $I_{e}$. Specifically, for the visual feature $F_{v}$, we use the trigger projector (see \ref{sec:Trigger Projector}) to adjust its dimension from $F_{v}$(1 $\times$ 768 each) to $F_{v}$(1 $\times$ 4096 each) so that it can be processed by LLMs. We then concatenate($\oplus$) the visual features $F_{v}$ and the language prompt 1 $LP_{1}$, and feed them into LLMs to generate entity-related information $I_{e}$ using greedy search. This can be expressed as:
\begin{equation}
	\begin{aligned}
		& I_{e} = \text{LLM}(F_{v} \oplus LP_{1}).
	\end{aligned}
\end{equation}
The design of language prompt 1 is similar to that of EVCap~\cite{li2024evcap} and Minigpt-4~\cite{zhu2023minigpt}:

\noindent
\begin{minipage}{0.99\columnwidth}
	\centering
	\vspace{0.5mm}
	\begin{tcolorbox}
			\raggedright
			\texttt{\#\#\#Human:} \texttt{<Img><ProjFeature></Img>} \texttt{What are they?} 
			\texttt{\#\#\#Assistant:}
		\end{tcolorbox}
	\vspace{0.5mm}
\end{minipage}
where $ProjFeature$ denotes the concatenation of $F_{v}$.

\subsection{Multi-Modal Purification Module}
In line with recent studies~\cite{shi2023large, fang2024enhancing, xu2024unsupervised}, we find that information retrieved through retrieval does not always play a beneficial role in generation, it may also introduce additional noise. To solve this problem, we consider adding an information filter after trigger-augmented module 1 to filter the noise in the generated information and produce more accurate entity information. Specifically, we introduce a multi-modal purification module (See Fig.~\ref{fig_2}(a)), based on the frozen Q-Former~\cite{li2023blip}. The multi-modal purification module requires three inputs: entity-related information, visual features, and entity query tokens. Among them, the entity-related information generated by trigger-augmented module 1 is of variable length, so we use learnable entity query tokens to extract fixed-length entity features. At the same time, because the entity-related information generated by trigger-augmented module 1 may contain noise unrelated to the entities in the image, visual features are incorporated to retain entity-related information while removing redundant noise. Overall, multi-modal purification employs learnable entity-query tokens along with visual features to compress, purify, and refine entity-related information, thereby generating entity features. This process can be expressed as:
\begin{equation}
	\begin{aligned}
		& F_{e} = \Omega(I_{e}, F_{v}, T_{e}),
	\end{aligned}
\end{equation}
where $F_{e}$ represents entity features, $\Omega(\cdot)$ denotes multi-modal purification, $I_{e}$ refers entity-related information, $F_{v}$ stands for visual features, and $T_{e}$ represents entity query tokens.

\subsection{Trigger Projector}
\label{sec:Trigger Projector}
To enable LLMs to understand the image content, we need to align the visual and textual features. Unlike BLIP2~\cite{li2023blip}, LLaVA~\cite{liu2024visual}, and EVCap~\cite{li2024evcap}, we select a frozen projector with superior performance and use another projector to trigger it. This approach not only reduces the number of trainable parameters but also shortens the training time. Moreover, fine-tuning the model on a different dataset with the frozen projector can alter the learned data distribution, enhancing model robustness. 
Specifically, our trigger projector consists of a trainable linear projector and a frozen pre-trained projector, as shown in Fig.~\ref{fig_1}. To maintain strong alignment capability, we select the LLaVA\cite{liu2024visual} pre-trained projector as our frozen pre-trained projector and construct a new learnable linear projector as the trigger. Our learnable linear projector first projects 768-dimensional features to 1024 dimensions, serving as the trigger for freezing the projector. The frozen pre-trained projector then maps the features to 4096 dimensions to align the visual and textual representations. This process can be expressed as:
\begin{equation}
	\begin{aligned}
			& F^{(b, l, 768)} \to F^{(b, l, 1024)}, \\
			& F^{(b, l, 1024)} \to F^{(b, l, 4096)},
		\end{aligned}
\end{equation}
where $F$ is input features, $b$ is batch size of input images, and $l$ is sequence length of the features.

\subsection{Trigger-Augmented Module 2}
To enable visual and entity features to generate image captions, we introduce trigger-augmented module 2 (TA2), a decoding architecture that concatenates these features, utilizes the trigger projector to map them to dimensions compatible with LLMs, and then concatenates language prompt 2 as input to the LLM decoder. This can be expressed as:
\begin{equation}
	\begin{aligned}
		& \text{Output} = \text{LLM}(F_{v} \oplus F_{e} \oplus LP_{2}),
	\end{aligned}
\end{equation}
where $F_{v}$ represents visual features, $F_{e}$ represents entity features, $LP_{2}$ represents language prompt 2. The design of language prompt 2 is follows:

\noindent
\begin{minipage}{0.99\columnwidth}
	\centering
	\vspace{1mm}
	\begin{tcolorbox}
			\raggedright
			\texttt{\#\#\#Human:} \texttt{<Img><ProjFeature></Img>} \texttt{Describe this image in detail.} 
			\texttt{\#\#\#Assistant:}
		\end{tcolorbox}
	\vspace{0.5mm}
\end{minipage}
where $ProjFeature$ denotes the concatenation of $F_{v}$, and $F_{e}$.

It should be noted that trigger-augmented module 1 and trigger-augmented module 2 are almost identical in structure, with the only difference being that the language prompts used for different inputs vary.

\subsection{Training and Inference}
In training phase, for a given image-text pair, our model extracts visual features from the image and generates entity features, combining them with language prompt 2 to form an embedding prompt of length $n$, denoted as $\{w_{i}\}_{i=1}^{n}$. The text is then converted into a collection of words of length $l$, denoted as $\{c_{i}\}_{i=1}^{l}$. The concatenation of $\{w_{i}\}_{i=1}^{n}$ and $\{c_{i}\}_{i=1}^{l}$ is used as the input to the LLM's decoder, which predicts each word in an autoregressive manner until a set maximum length is reached or a terminal token is encountered. We use cross-entropy loss as the objective function to optimize our model, which can be expressed as follows:
\begin{equation}
	\begin{aligned}
		& L_{\theta} = -\sum_{i=1}^{l} \log p_{\theta} (c_i \mid w_1, \dots, w_n, c_1, \dots, c_{i-1}),
	\end{aligned}
\end{equation}
where $\theta$ represents the trainable parameters of TPCap, $l$ represents the current sentence length, ${w_{1}, \dots, w_{n}}$ represents the embedding prompt of length $n$, and $c_{i}$ represents the current predicted word.

In inference phase, we only require images as input. Our model extracts features from images, generates entity features through trigger-augmented module 1 with multi-modal purification, and then concatenates visual features, entity features, and language prompt 2 as input to the LLM decoder to generate an image description.

\begin{table*}[t]
\centering
\caption{Compared with other popular models on MSCOCO, NoCaps, and Flickr30k, we also compare the number of trainable parameters and present the training data used. Where B@4, M, C, and S represent BLEU@4, METEOR, CIDEr, and SPICE, respectively. * denotes using a memory bank, {\dag}  denotes the result we reproduced on a single RTX 4090 GPU, the same device used for our method.} 
\vspace*{-0.5\baselineskip}	
\label{tab:overall}
\resizebox{\linewidth}{!}{
\begin{tabular}{l|cc|cccc|cccccccc|cc}
\toprule[1pt]
\multirow{3}{*}{\textbf{Method}} & \multicolumn{2}{c|}{\textbf{Training}} & \multicolumn{4}{c|}{\textbf{COCO}} & \multicolumn{8}{c|}{\textbf{NoCaps Val}} & \multicolumn{2}{c}{\textbf{Flickr30k}} \\
& \multirow{2}{*}{Data} & \multirow{2}{*}{Para.}
& \multicolumn{4}{c|}{Test}
& \multicolumn{2}{c|}{In-domain} & \multicolumn{2}{c|}{Near-domain} 
& \multicolumn{2}{c|}{Out-domain} & \multicolumn{2}{c|}{Overall}
& \multicolumn{2}{c}{Test} \\
& & 
& \multicolumn{1}{c}{B@4} & \multicolumn{1}{c}{M} & \multicolumn{1}{c}{C} & \multicolumn{1}{c|}{S}
& \multicolumn{1}{c}{C} & \multicolumn{1}{c|}{S} 
& \multicolumn{1}{c}{C} & \multicolumn{1}{c|}{S} 
& \multicolumn{1}{c}{C} & \multicolumn{1}{c|}{S} 
& \multicolumn{1}{c}{C} & \multicolumn{1}{c|}{S} 
& \multicolumn{1}{c}{C} & \multicolumn{1}{c}{S} \\
\midrule
\multicolumn{17}{l}{\demph{\textbf{Heavyweight-Training Models}}} \\
\demph{VinVL~\cite{zhang2021vinvl}} & \demph{8.9M}  & \demph{110M}  &\demph{38.2} &\demph{30.3} &\demph{129.3} & \demph{23.6}  & \demph{96.8} & \demph{13.5} & \demph{90.7} & \demph{13.1} & \demph{87.4} & \demph{11.6} & \demph{90.9} & \demph{12.8} & \demph{--} & \demph{--}  \\
\demph{AoANet+MA*~\cite{fei2021memory}} & \demph{COCO}  & \demph{--}  & \demph{38.0} & \demph{28.7} & \demph{121.0}  & \demph{21.8} & \demph{--}  & \demph{--} & \demph{--}  & \demph{--}  & \demph{--} & \demph{--} & \demph{--}  & \demph{--} & \demph{--} & \demph{--}   \\
\demph{NOC-REK*~\cite{vo2022noc}} & \demph{COCO} & \demph{110M} & \demph{--} & \demph{--} & \demph{--}  & \demph{--}  & \demph{104.7} & \demph{14.8} & \demph{100.2} & \demph{14.1} & \demph{100.7} & \demph{13.0} & \demph{100.9} & \demph{14.0}  & \demph{--} & \demph{--}  \\
\demph{RCA-NOC*~\cite{fan2023rca}}  & \demph{COCO} & \demph{110M} & \demph{37.4} & \demph{29.6} & \demph{128.4}  & \demph{23.1}  & \demph{92.2} & \demph{12.9} & \demph{87.8} & \demph{12.6} & \demph{87.5} & \demph{11.5} & \demph{88.3} & \demph{12.4} & \demph{--} & \demph{--} \\
\demph{ViECap \demph{$_\text{GPT2}$}~\cite{fei2023transferable}} & \demph{COCO}   & \demph{124M} & \demph{27.2} & \demph{24.8} & \demph{92.9}  & \demph{18.2}  & \demph{61.1} & \demph{10.4} & \demph{64.3} & \demph{9.9} & \demph{65.0} & \demph{8.6} & \demph{66.2} & \demph{9.5}  & \demph{47.9} & \demph{13.6} \\
\demph{InstructBLIP \demph{$_\text{Vicuna-13B}$}~\cite{NEURIPS2023_9a6a435e}} & \demph{129M}  & \demph{188M} & \demph{--} & \demph{--} & \demph{--}  & \demph{--}  & \demph{--}  & \demph{--}  & \demph{--}  & \demph{--} & \demph{--}  & \demph{--}  & \demph{121.9}  & \demph{--}  & \demph{82.8} & \demph{--} \\
\demph{OSCAR~\cite{li2020oscar}} & \demph{4.1M}  & \demph{338M}  & \demph{37.4} & \demph{30.7} & \demph{127.8} & \demph{23.5} & \demph{83.4} & \demph{12.0} & \demph{81.6} & \demph{12.0} & \demph{77.6} & \demph{10.6} & \demph{81.1} & \demph{11.7} & \demph{--} & \demph{--} \\
\demph{BLIP~\cite{li2022blip}}  & \demph{129M}  & \demph{446M} & \demph{40.4} & \demph{--} & \demph{136.7} & \demph{--} & \demph{114.9} & \demph{15.2} & \demph{112.1} & \demph{14.9} & \demph{115.3} & \demph{14.4} & \demph{113.2} & \demph{14.8} & \demph{--} & \demph{--} \\
\demph{BLIP-2 \demph{$_\text{FlanT5-XL}$}~\cite{li2023blip}}  & \demph{129M}  & \demph{1.2B} & \demph{42.4} & \demph{--} & \demph{144.5} & \demph{--} & \demph{123.7} & \demph{16.3} & \demph{120.2} & \demph{15.9} & \demph{124.8} & \demph{15.1} & \demph{121.6} & \demph{15.8}  & \demph{--} & \demph{--} \\
\demph{REVEAL* \demph{$_\text{T5}$}~\cite{hu2023reveal}}  & \demph{1.3B}  & \demph{2.1B}  & \demph{--} & \demph{--} & \demph{145.4} & \demph{--}  & \demph{--}  & \demph{--}  & \demph{--}  & \demph{--} & \demph{--}  & \demph{--}  & \demph{123.0} & \demph{--}  & \demph{--} & \demph{--} \\
\midrule
\multicolumn{17}{l}{\textbf{Lightweight-Training Models}} \\
MiniGPT4 \demph{$_\text{Vicuna-13B}$}~\cite{zhu2023minigpt} & 5M  & 3.94M & 38.0 & 29.6 & 129.6  & 23.4  & 99.0  & 14.8  & 106.9  & 15.3 & 110.8  & \textbf{14.9}  & 108.8 & \textbf{15.1}  & 78.4 & 16.9 \\
SmallCap* \demph{$_\text{GPT2}$}~\cite{ramos2023smallcap} & COCO  & 7M & 37.0 & 27.9 & 119.7  & 21.3  & --  & --  & --  & -- & --  & --  & -- & --  & 60.6 & -- \\
ClipCap \demph{$_\text{GPT2}$}~\cite{mokady2021clipcap}  & COCO  & 43M & 33.5 & 27.5 & 113.1  & 21.1   & 84.9 & 12.1 & 66.8 & 10.9 & 49.1 & 9.6 & 65.8 & 10.9    & -- & -- \\
EVCap*\dag \demph{$_\text{Vicuna-7B}$}~\cite{li2024evcap} & COCO & 3.97M & 40.3 & \textbf{30.9} & 137.2  & 24.5  & 109.0 & 14.8 & 115.4 & 15.1 & 112.1 & 14.7 & 115.3 & 15.0 & 80.3 & 17.6 \\
\rowcolor{Gray}
Ours \demph{$_\text{Vicuna-7B}$} & COCO & \textbf{0.82M} & \textbf{40.7} & 30.6 & \textbf{138.6} & \textbf{24.6} & \textbf{113.9} & \textbf{15.3} & \textbf{118.5} & \textbf{15.5} & \textbf{114.4} & 14.4 & \textbf{118.2} & \textbf{15.1} & \textbf{84.3} & \textbf{18.2} \\
\midrule
\multicolumn{17}{l}{\demph{\textbf{Specialist SOTAs}}} \\
\demph{Qwen-VL \demph{$_\text{Qwen-7B}$}~\cite{bai2023qwen}} & \demph{1.4B} & \demph{9.6B} & \demph{--} & \demph{--} & \demph{--}  & \demph{--}  & \demph{--}  & \demph{--}  & \demph{--}  & \demph{--} & \demph{--}  & \demph{--}  & \demph{121.4} & \demph{--} & \demph{85.8} & \demph{--} \\
\demph{CogVLM \demph{$_\text{Vicuna-7B}$}~\cite{wang2023cogvlm}} & \demph{1.5B}  & \demph{6.5B} & \demph{--} & \demph{--} & \demph{148.7}  & \demph{--}  & \demph{--}  & \demph{--}  & \demph{--}  & \demph{--} & \demph{132.6}  & \demph{--}  & \demph{128.3} & \demph{--} & \demph{94.9} & \demph{--} \\
\demph{PaLI \demph{$_\text{mT5-XXL}$}~\cite{chen2022pali}} & \demph{1.6B}  & \demph{17B} & \demph{--} & \demph{--} & \demph{149.1}  & \demph{--}  & \demph{--}  & \demph{--}  & \demph{--}  & \demph{--} & \demph{--}  & \demph{--}  & \demph{127.0}  & \demph{--} & \demph{--} & \demph{--}\\
\demph{PaLI-X \demph{$_\text{UL2-32B}$}~\cite{chen2023pali}} & \demph{2.2B}  & \demph{55B} & \demph{--} & \demph{--} & \demph{149.2}  & \demph{--}  & \demph{--}  & \demph{--}  & \demph{--}  & \demph{--} & \demph{--}  & \demph{--}  & \demph{126.3}  & \demph{--} & \demph{--} & \demph{--} \\
\bottomrule[1pt]
\end{tabular}
}
\end{table*}

\section{Experiment and Analysis}
To validate the proposed TPCap method's superiority, it is compared with multiple state-of-the-art image captioning approaches on four large-scale datasets, namely, COCO, NoCaps, Flickr30k, and WHOOPS.

\subsection{Datasets and Evaluation Metrics}
\textbf{COCO.}
COCO~\cite{lin2014microsoft} contains 80 different object categories. It is mainly used for object recognition, object detection, image segmentation, and image captioning tasks. At first, COCO contained 164,062 images, including 82,783 images in the training set, 40,504 images in the validation set, and 40,775 images in the test set. Karpathy then repartitioned the COCO dataset~\cite{karpathy2015deep} by taking 10,000 images from the validation set, assigning 5,000 to the new validation set, 5,000 to the test set, and the remaining 30,504 images to the training set, which, together with the original 82,783 images, constituted the new training set.

\textbf{NoCaps.}
In order for the model to learn a larger variety of visual concepts, it is better to learn with less supervision, NoCaps~\cite{agrawal2019nocaps} selects 4,500 images from the 41,620 validation set of Open Images as the validation set, and 10,600 images from the 125,436 test set of Open Images as the test set, with an average of 4.0 object classes and 8.0 object instances per image. NoCaps also divides the data into in-domain, near-domain, and out-of-domain, where the in-domain contains only COCO classes, the near-domain contains both COCO classes and novel classes, and the out-of-domain contains only the novel class.

\textbf{Flickr30k.}
Flickr30k~\cite{young2014image} consists of 31,783 images related to everyday activities, events, and scenes, along with their corresponding 158,915 captions. It is an extension of Flickr8k. The images were taken from Flickr, a well-known online image-sharing platform. Different from COCO, Flickr30k doesn't specify the number of classes. At the same time, the style of Flickr30k is more concise and direct than COCO. Currently, working on Flickr30K usually follows the practice of Karpathy split, where 1,000 images are used for validation, 1,000 images are used for testing, and the remaining images are used for training.

\textbf{WHOOPS.}
WHOOPS~\cite{bitton2023breaking} contains 500 synthetic images and 10,874 annotations, the images created by designers using publicly available image generation tools such as MidJourney. These images deliberately violate common sense and are designed to challenge the AI model's understanding of common sense and compositionality. WHOOPS The application area of the dataset is mainly focused on improving the visual commonsense reasoning ability of AI models, especially when dealing with atypical and illogical image scenes. Different from other datasets, WHOOPS only has a test set without a training or validation set.

\textbf{Metrics.}
To evaluate our model and compare it with other methods, we use common evaluation metrics: BLEU (B@1$\sim$4)~\cite{papineni2002bleu}, METEOR (M)~\cite{banerjee2005meteor}, ROUGE (R)~\cite{lin2004rouge}, CIDEr (C)~\cite{vedantam2015cider}, and SPICE (S)~\cite{anderson2016spice} to assess the quality of image captioning. The BLEU is simple to compute and suitable for multilingual evaluation. The METEOR takes into account lexical richness, syntax, and semantics, providing a more flexible evaluation than BLEU. The ROUGE emphasizes recall and is suitable for capturing more relevant information. The CIDEr is designed for image captioning tasks and is able to assess description quality more accurately by considering multiple reference descriptions. The SPICE, on the other hand, focuses on evaluation at the semantic level, providing a more nuanced comparison of content than lexical matching. The metrics currently considered the most important are CIDEr and SPICE.

\subsection{Implementation Details}
In TPCap, we utilize EVA-CLIP-g~\footnote{https://github.com/baaivision/EVA} as the vision encoder, which outputs image features of size ($257 \times 1408$). Then, to reduce computational overhead and obtain more accurate visual features, we use a frozen Q-Former~\cite{li2023blip} with 32 learnable image query tokens to compress the image features, which output image features of size ($32 \times 768$). Then, the image features input into trigger-augmented module 1 to generate variable-length entity-related information under the greedy search. Then, the entity-related information is input into the multi-modal purification. We use 8 entity query tokens, combined with visual features, to compress, purify, and refine them, and finally generate the entity features. Then, the visual features and entity features are concatenated and input into trigger-augmented module 2 to generate image captioning. Vicuna-7b-v1.3~\cite{zheng2023judging} is chosen for our large language model, and the maximum text length is set to 160. The parameters of the frozen Q-Former and the frozen multi-modal purification are from BLIP-2~\cite{li2023blip}, and the parameters of the frozen projector in the trigger projector are from LLaVA~\cite{liu2024visual}. We trained on COCO for one epoch. The initial learning rate is set to $1e^{-4}$, and the LinearWarmupCosineLRScheduler is used to manage the learning rate decay, with a minimum value of $8e^{-5}$. The entire training process is carried out on a single NVIDIA GeForce RTX 4090 GPU, using mixed precision.

\subsection{Comparison with State-of-the-Art Methods}
In this section, we compare our method with other state-of-the-art (SOTA) models on COCO, NoCaps, Flickr30k, and WHOOPS.

\subsubsection{Comparisons on COCO, NoCaps, and Flickr30k}
We compare our method with the most popular heavyweight models, lightweight models, and specialized SOTAs on classic test datasets: COCO test, NoCaps val, and Flickr30k test. The comparison results are presented in Table.~\ref{tab:overall}, demonstrate that the proposed lightweight method is superior to many state-of-the-art heavyweight methods and outperforms all the lightweight methods. 

\textbf{Compared to the lightweight-training models.} 
Our model is trained only on COCO, and the number of training parameters in our model is the smallest among all the compared models, with only 0.82M parameters, which is 1/5 of the number of parameters in the second smallest, MiniGPT-4, at 3.94M. Despite using the least amount of training data and the fewest number of parameters, we still outperform all the lightweight models by an average of 13.7 on the COCO test, 21.6 on the NoCaps validation set, and 11.2 on the Flickr30k test, based on the main CIDEr scores. This highlights the efficient performance of our method while maintaining its effectiveness.

\textbf{Compared to the heavyweight models.} Our performance is higher than most, particularly outperforming all models trained on COCO, which proves that our method is effective. We are only below InstructBLIP, BLIP-2, and REVEAL. Our CIDEr score is 6.8 lower than the highest on the COCO test, and our CIDEr score on the NoCaps validation overall case is 4.8 lower than the highest, while our SPICE score is 0.7 lower than the highest. However, it should be noted that the LLM we use is Vicuna-7B, which has only 7B parameters, the smallest among all the models. At the same time, our trainable parameters are also the fewest among all the models, making our results acceptable.

\textbf{Compared to the specialist models.} We are nearly 10 points below the top CIDEr scores on all datasets. However, our training dataset is approximately 1/13750 of the largest dataset and 1/8750 of the smallest dataset. Additionally, the number of trainable parameters in our model is 1/67073 of the maximum and 1/7927 of the minimum. Given these factors, our results are acceptable.

\textbf{Comparison with other methods using RAG.} The experimental results show that, except for the REVEAL model, the performance of the proposed model is higher than that of other models using RAG, which proves that our special retrieval enhancement method based on the trigger-augmented module is effective. At the same time, our method does not require an additional memory bank, and our trainable parameters are the fewest, proving that our method is both concise and efficient.

\subsubsection{Comparisons on WHOOPS}


\begin{table}[!t]
\centering
\caption{The results tested on the commonsense-violation dataset WHOOPS show that our model demonstrates a strong open-world understanding ability.}
\label{tab:whoops}
\vspace*{-0.5\baselineskip}	
\resizebox{0.8\linewidth}{!}{
\begin{tabular}{lcccc}
\toprule[1pt]
\multirow{1}{*}{\textbf{Method}} 
& B@4 & M & C & S\\ 
\midrule
\multicolumn{5}{l}{\emph{Only Pre-Trained Models}} \\
BLIP ~\cite{bitton2023breaking} & 13  & -- & 65 & -- \\ 
BLIP-2 \demph{$_\text{FlanT5-XXL}$}~\cite{li2024evcap}  & 28  & 26.7 & 93.1 & 17.9 \\ 
\midrule
\multicolumn{5}{l}{\emph{Finetuned Models on COCO}} \\
MiniGPT4~\cite{zhu2023minigpt}  & 24.2   & 26.7  & 84.8 & 18.2 \\ 
BLIP ~\cite{li2022blip} & 22.9  & 25.0 & 79.3 & 17.1 \\ 
BLIP-2 \demph{$_\text{FlanT5-XL}$}~\cite{li2023blip} & 25.8  & 27.0 & 89.1 & 18.3 \\ 
EVCap \demph{$_\text{Vicuna-13B}$} ~\cite{li2024evcap} & 24.1   & 26.1 & 85.3 & 17.7 \\ 
\rowcolor{Gray}
Ours \demph{$_\text{Vicuna-7B}$} & \textbf{28.1} & \textbf{27.6} & \textbf{121.8} & \textbf{21.3} \\
\bottomrule[1pt]
\end{tabular}
}
\end{table}

To validate the model's ability to describe the open world, we compare TPCap with several recent models on the WHOOPS dataset, which contains images that violate common sense. The comparison results are presented in Table.~\ref{tab:whoops}. The TPCap have achieved a score of 28.1 on BLEU4, 27.6 on METEOR, 121.8 on CIDEr, and 21.3 on SPICE, exceeding the second-place model by 28.7 and 3.1 on the main CIDEr and SPICE, respectively. As shown in Table.~\ref{tab:whoops}, our method outperforms all methods retraining cross-modal projectors on COCO by using learnable projectors to trigger frozen projectors, which demonstrates that trigger-projectors based on frozen projectors are highly effective.  Our method even outperforms some pre-trained large models, demonstrating the superiority of our approach. At the same time, there are a large number of samples in the WHOOPS dataset that do not appear in COCO, and our high performance further proves the zero-shot capability of our method. 

\subsection{Ablation Studies and Analysis}

In what follows,  the proposed TPCap method is comprehensively analyzed from 3 aspects to investigate the logic behind its superiority.

\subsubsection{Role of Trigger-Augmented Module}

\begin{table}[t]

	\centering
	\caption{For the ablation experiments of trigger augmented modules and the multi-modal purification, where TA 1 refers to trigger-augmented module 1, TA 2 refers to trigger-augmented module 2, and MP refers to the multi-modal purification.} 
	\vspace*{-0.5\baselineskip}	
	\label{ablation}
	\resizebox{\linewidth}{!}{
		\begin{threeparttable}
		\begin{tabular}{c|c|c|cc|cc|cc}
			\toprule[1pt]
			\multirow{2}{*}{\textbf{TA 2}} &
			\multirow{2}{*}{\textbf{TA 1}} &
			\multirow{2}{*}{MP} & \multicolumn{2}{c|}{\textbf{COCO}} & \multicolumn{2}{c|}{\textbf{NoCaps}} & \multicolumn{2}{c}{\textbf{Flickr30k}} \\
			& & & \multicolumn{1}{c}{C} & \multicolumn{1}{c|}{S}
			& \multicolumn{1}{c}{C} & \multicolumn{1}{c|}{S} 
			& \multicolumn{1}{c}{C} & \multicolumn{1}{c}{S}  \\
			\midrule
			+ & + & + & \textbf{138.6} & \textbf{24.6} & \textbf{118.2} & \textbf{15.1} & \textbf{84.3} & \textbf{18.2} \\
			+ & + & - & 135.2 & 24.2 & 111.0 & 14.5 & 78.7 & 17.2 \\
			+ & - & + & 136.7 & 24.2 & 116.5 & 15.0 & 83.6 & 18.0 \\
			+ & - & - & 136.2 & 24.3 & 115.4 & 14.8 & 82.2 & 17.8 \\
			\bottomrule[1pt]
		\end{tabular}
	\begin{tablenotes}[para, flushleft]
		\item \textbf{Note:} As mentioned, using only trigger-augmented module 1 to generate relevant information introduces noise and leads to worse performance than the baseline. A comparison between line 2 and line 4 shows this.
	\end{tablenotes}
	\end{threeparttable}
}
\end{table}
 
There are two trigger-augmented modules in our model. The trigger-augmented module 1 is used to generate entity-related information, which is a special retrieval generation method. The trigger-augmented module 2 is used to generate the final image captions (only the model with trigger-augmented module 2 was used as our baseline). In particular, trigger-augmented module 1 is a very important part of our model. We use different language prompts from trigger-augmented module 2 to guide trigger-augmented module 1 in generating more information related to entities, which effectively ensures the diversity of entity descriptions in our model. In order to verify the effectiveness of trigger-augmented module 1, we conducted ablation experiments on it, and the results are shown in Table.~\ref{ablation}. After removing trigger-augmented module 1, the CIDEr scores of the model on all test datasets significantly decreased. However, the decrease in the SPCIE index was not obvious, indicating that the accuracy of the model for entity relationship descriptions did not decrease significantly, but the accuracy of entity descriptions decreased greatly. This proves the effectiveness of our trigger-augmented module 1.

\subsubsection{Influence of Multi-Modal Purification}
\begin{figure}[!t]
	\centering
	\includegraphics[width=0.5\textwidth]{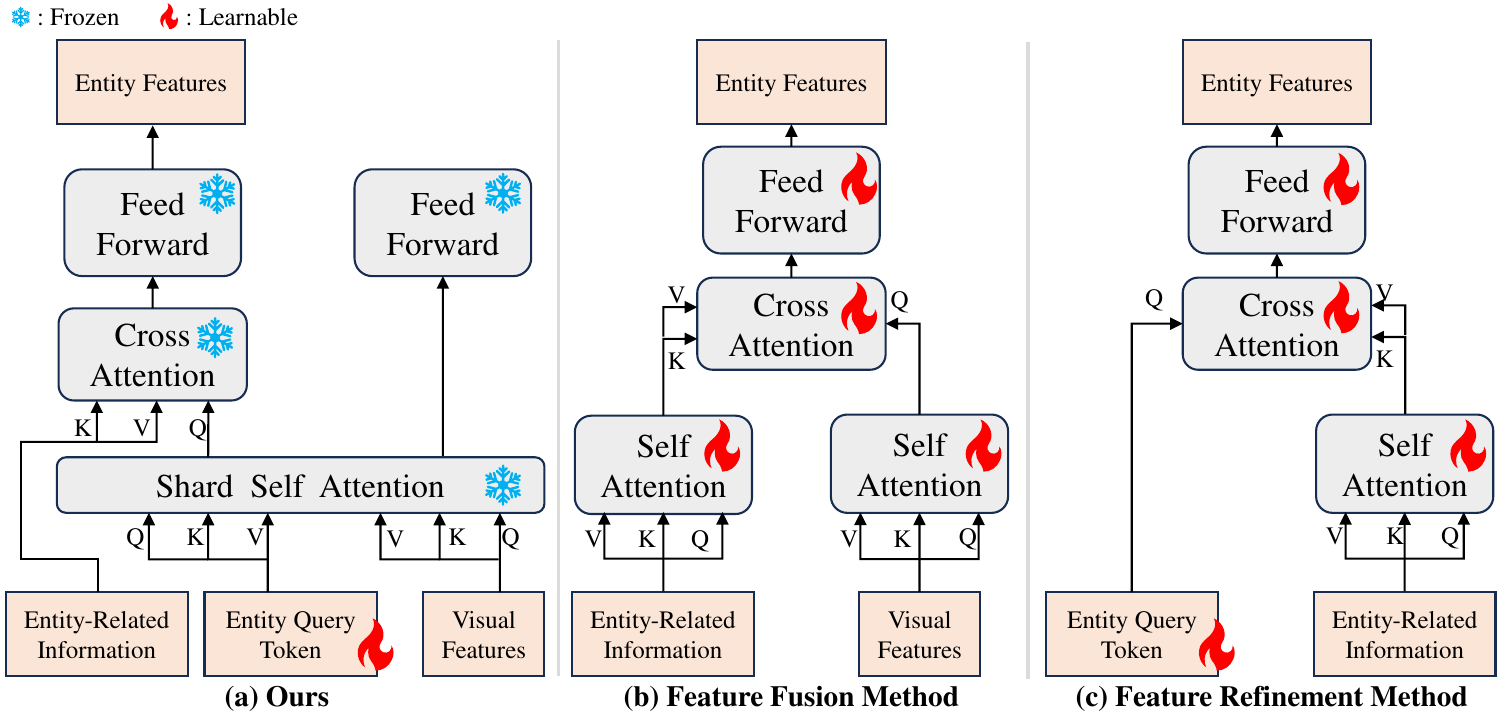}
	\caption{Three different types of entity-related information processing methods: \textbf{(a) ours} is used to compress, purify, and refine the entity-related information; \textbf{(b) feature fusion method} using cross-attention achieve information fusion; \textbf{(c) feature refinement method} using cross-attention and learnable tokens, which refines the features. \textbf{Note}, for simplicity, we omit the variation in feature dimensions through the linear layer.}
	\label{fig_2}
\end{figure}
To solve the problem that the possibly noisy entity-related information generated by trigger-augmented module 1 may mislead LLMs into generating incorrect information, we consider sanitizing the entity-related information. We design three methods to purify the entity-related information, as illustrated in Fig.~\ref{fig_2}. 

\textbf{Multi-modal purification (ours).} We first introduce multi-modal purification, which, similar to the customized Q-Former~\cite{li2024evcap}, is a frozen module with parameters derived from the Q-Former~\cite{li2023blip}. With parameter-shared self-attention, learnable entity query tokens are able to capture visual features and contextual relationships. Then, the entity query tokens interact with the entity-related information through cross-attention to compress the information and remove redundant noise, converting it into entity features. 
\textbf{Feature fusion method.}  Through two learnable self-attentions, key information is obtained by capturing the contextual relationships of entity-related information and visual features, respectively. Then, visual features are used as the query, and entity features are obtained through learnable cross-attention fusion. 
\textbf{Feature refinement method.} The learnable entity query token is used as a query to refine entity-related information through cross-attention, obtaining key information from it. 

\textbf{Analysis of results.} As shown in Table.~\ref{ablation_multi_modal}. Our multi-modal purification achieves the best performance, which is due to the shared self-attention that enables the learnable entity-query token to capture the context of visual features. This plays a key role in alignment and reduces noise when fused with entity-related information. Additionally, a fixed number of entity-query tokens plays a crucial role in compressing and refining the entity-related information. The reason for the poor performance of the feature fusion method is that the entity features are obtained only through cross-attention information fusion, without considering further processing of the noise in the entity-related information. The reason for the poor performance of the feature refinement method is that the visual features are not integrated, and only the entity-related information is refined. This makes the model too sensitive to the training data, resulting in good performance on the COCO with the same distribution, but poor performance on Flickr30k, which has a different distribution.

\begin{table}[t]
	
	\centering
	\caption{Comparative experiments of different refinement methods.} 
	\vspace*{-0.5\baselineskip}	
	\label{ablation_multi_modal}
	\resizebox{\linewidth}{!}{
			\begin{tabular}{l|cc|cc|cc}
				\toprule[1pt]
				\multirow{2}{*}{Method} & \multicolumn{2}{c|}{\textbf{COCO}} & \multicolumn{2}{c|}{\textbf{NoCaps}} & \multicolumn{2}{c}{\textbf{Flickr30k}} \\
				& \multicolumn{1}{c}{C} & \multicolumn{1}{c|}{S}
				& \multicolumn{1}{c}{C} & \multicolumn{1}{c|}{S} 
				& \multicolumn{1}{c}{C} & \multicolumn{1}{c}{S}  \\
				\midrule
				Multi-Modal Purification & \textbf{138.6} & \textbf{24.6} & \textbf{118.2} & \textbf{15.1} & \textbf{84.3} & \textbf{18.2} \\
				Feature Fusion &  135.8 & 24.3 & 113.4 & 14.6 & 79.1 & 17.2 \\
				Feature Refinement &  137.4 & 24.4 & 114.0 & 14.8 & 78.8 & 17.2 \\
				\bottomrule[1pt]
			\end{tabular}
	}
\end{table}

\subsubsection{Impact of Trigger Projector}
\begin{figure}[!t]
	\centering
	\includegraphics[width=0.5\textwidth]{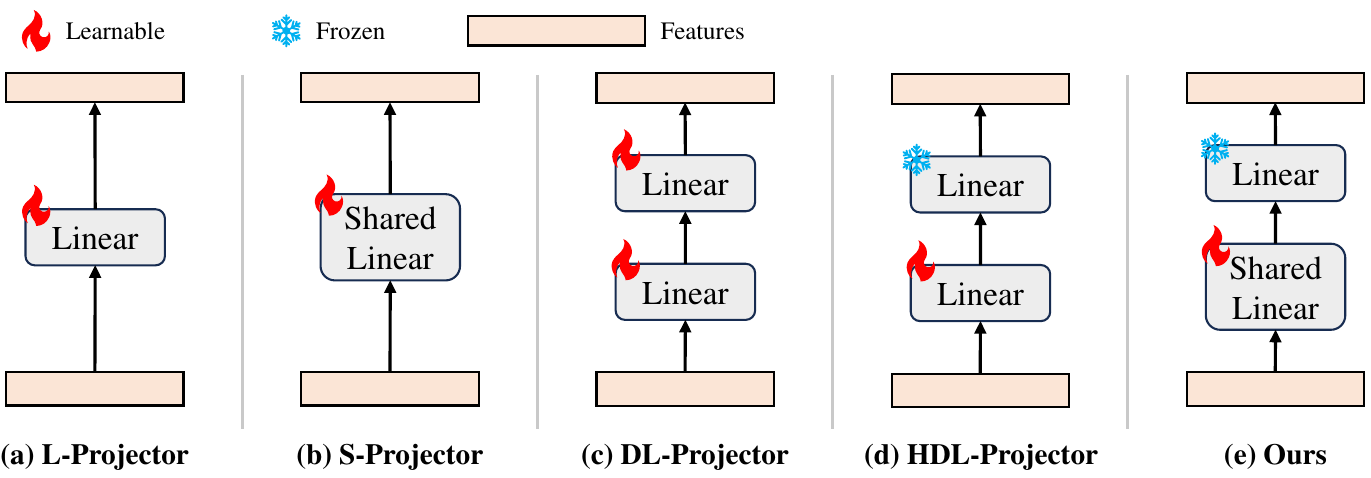}
	\caption{Five different projector types. \textbf{(a) L-Projector} consists of a single linear layer that projects the input features dimensions to 4096; \textbf{(b) S-Projector} consists of a single shared linear layer, indicating that the projector parameters are shared between trigger-augmented module 1 and trigger-augmented module 2; \textbf{(c) DL-Projector} consists of two linear layers: the first linear layer projects the input features from 768 to 1024, and the second linear layer projects the dimensions from 1024 to 4096; \textbf{(d) HDL-Projector} consists of a learnable linear layer that projects the input feature dimensions to 1024 and a frozen linear layer that projects the feature dimensions from 1024 to 4096; \textbf{(e) Ours} and e are similar, except that the parameters of the first learnable linear layer are shared. \textbf{Note} that the frozen linear layer requires an input dimension of 1024, while our input feature dimension is fixed at 768, so the frozen linear layer cannot exist alone.}
	\label{fig_3}
\end{figure}

To verify the effectiveness of the trigger projector, we designed five different projector types, as shown in Fig.~\ref{fig_3}. We incorporated each of these projector types into the trigger-augmented module for experiments, and the results are shown in Table.~\ref{tab:trigger projector}. 

\textbf{Comparing (6) with (1) and (3) in Table.~\ref{tab:trigger projector}}. Respectively, the results of (6) outperform those of (1) and (3) on all datasets. At the same time, the number of parameters in (6) is the smallest, which demonstrates the superiority of our trigger projector. By comparing the three architectures, the importance of using the trigger projector can be highlighted. The L-Projector is highly sensitive to the distribution of fine-tuning data, resulting in poor generalization performance. The DL-Projector has a two-layer trainable linear, which makes visual features and text features difficult to align in fine-tuning. Our method uses a learnable linear layer to trigger a frozen linear layer. By fine-tuning the projection distribution based on the original projection distribution of the frozen linear layer using the COCO dataset, the model achieves better generalization ability while reducing the number of parameters in the learnable linear layer, facilitating convergence. 

\textbf{Comparing (1) and (2) in Table.~\ref{tab:trigger projector}}. The performance of (2) is slightly lower than that of (1), which proves that our trigger-augmented module 1 and trigger-augmented module 2 function separately under different language prompts. However, the small difference in performance indicates that the shared parameters are effective. 

\textbf{Comparing (6) with (4) and (5) in Table.~\ref{tab:trigger projector}}, the performance of (6) is the best, and the number of parameters is the least, which proves the effectiveness of our model. This is because, after fixing the projection space by freezing the linear layer, the parameter-sharing linear layer can better align the two features.

\begin{table*}[tb]
{\centering
\caption{Analysis of experiments for projector types in trigger-augmented module 1 (TA1) and trigger-augmented module 2 (TA2).}
\label{tab:trigger projector}
\vspace*{-0.5\baselineskip}	
\resizebox{\textwidth}{!}{
\begin{tabular}{c|c|c|c|cccc|cccccccc|cccc}
\toprule[1pt]
 &
\multirow{3}{*}{\textbf{TA1}} &
\multirow{3}{*}{\textbf{TA2}} &  
\multirow{3}{*}{\textbf{Para.}} &
\multicolumn{4}{c|}{\textbf{COCO}} & 
\multicolumn{8}{c|}{\textbf{NoCaps Val}} & \multicolumn{4}{c}{\textbf{Flickr30k}} \\ 
& & & 
& \multicolumn{4}{c|}{Test}
& \multicolumn{2}{c|}{In-domain} & \multicolumn{2}{c|}{Near-domain} 
& \multicolumn{2}{c|}{Out-domain} & \multicolumn{2}{c|}{Overall}
& \multicolumn{4}{c}{Test} \\
& &
& & \multicolumn{1}{c}{B@4} & \multicolumn{1}{c}{M} 
& \multicolumn{1}{c}{C} & \multicolumn{1}{c|}{S}
& \multicolumn{1}{c}{C} & \multicolumn{1}{c|}{S} 
& \multicolumn{1}{c}{C} & \multicolumn{1}{c|}{S} 
& \multicolumn{1}{c}{C} & \multicolumn{1}{c|}{S} 
& \multicolumn{1}{c}{C} & \multicolumn{1}{c|}{S}
& \multicolumn{1}{c}{B@4} & \multicolumn{1}{c}{M} 
& \multicolumn{1}{c}{C} & \multicolumn{1}{c}{S} \\
\midrule
(1) & L-Projector & L-Projector & 6.33M & 40.5 & \textbf{31.0} & 137.0 & 24.5 & 109.4 & 15.0 & 116.1 & 15.3 & 113.8 & 14.4 & 116.3 & \textbf{15.1} & 30.2 & 24.3 & 80.8 & 17.7 \\
(2) & S-Projector & S-Projector & 3.18M & 40.4 & 30.9 & 137.2 & 24.4 & 107.8 & 15.0 & 115.9 & 15.1 & 112.9 & 14.4 & 115.7 & 14.9 & 30.3 & 24.3 & 80.1 & 17.8 \\
(3) & DL-Projector & DL-Projector & 10.00M & 40.3 & 30.8 & 137.2 & 24.4 & 108.4 & 15.1 & 116.0 & 15.2 & 112.8 & 14.4 & 115.9 & 15.0 & 30.3 & 24.3 & 79.6 & 17.4 \\
(4) & L-Projector & HDL-Projector & 3.97M & 40.3 & 30.5 & 137.5 & 24.4 & 112.1 & \textbf{15.4} & 117.8 & 15.4 & 114.2 & \textbf{14.5} & 117.7 & \textbf{15.1} & 31.2 & 24.5 & 81.2 & 17.9 \\
(5) & HDL-Projector & L-Projector & 3.97M & 40.0 & 30.9 & 136.6 & 24.4 & 109.1 & 15.0 & 115.0 & 15.2 & 111.8 & \textbf{14.5} & 115.0 & 14.9 & 29.4 & 24.1 & 79.5 & 17.4 \\
(6) & Ours & Ours & \textbf{0.82M} & \textbf{40.7} & 30.6 & \textbf{138.6} & \textbf{24.6} & \textbf{113.9} & 15.3 & \textbf{118.5} & \textbf{15.5} & \textbf{114.4} & 14.4 & \textbf{118.2} & \textbf{15.1} & \textbf{32.2} & \textbf{24.9} & \textbf{84.3} & \textbf{18.2} \\
\bottomrule[1pt]
\end{tabular}
}
}
\end{table*}

\begin{figure}[!t]
	\centering
	\includegraphics[width=0.5\textwidth]{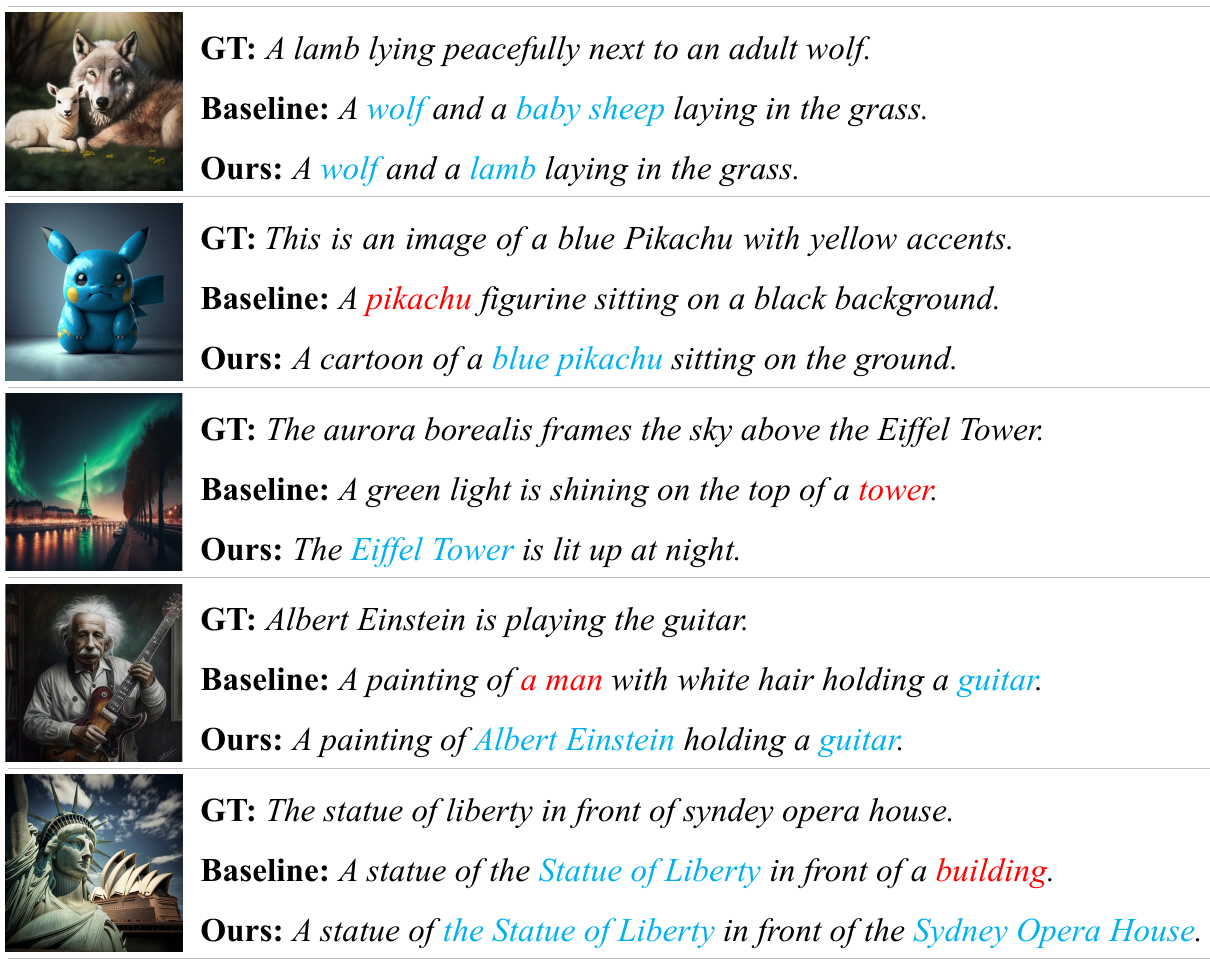}
	\caption{The results of WHOOPS show that our model has the ability to reason about commonsense compositionality and is capable of describing illogical images, demonstrating the ability to describe an open world. \textcolor[rgb]{0, 0.69, 0.94}{Blue} indicates entities. \textcolor{red}{Red} indicates entities whose descriptions are not accurate enough.}
	\label{fig_5}
\end{figure}

\subsection{Visualization} 
\subsubsection{Visualization on COCO, NoCaps, Flickr30k}
\begin{figure*}[!t]
	\centering
	\includegraphics[width=1\textwidth]{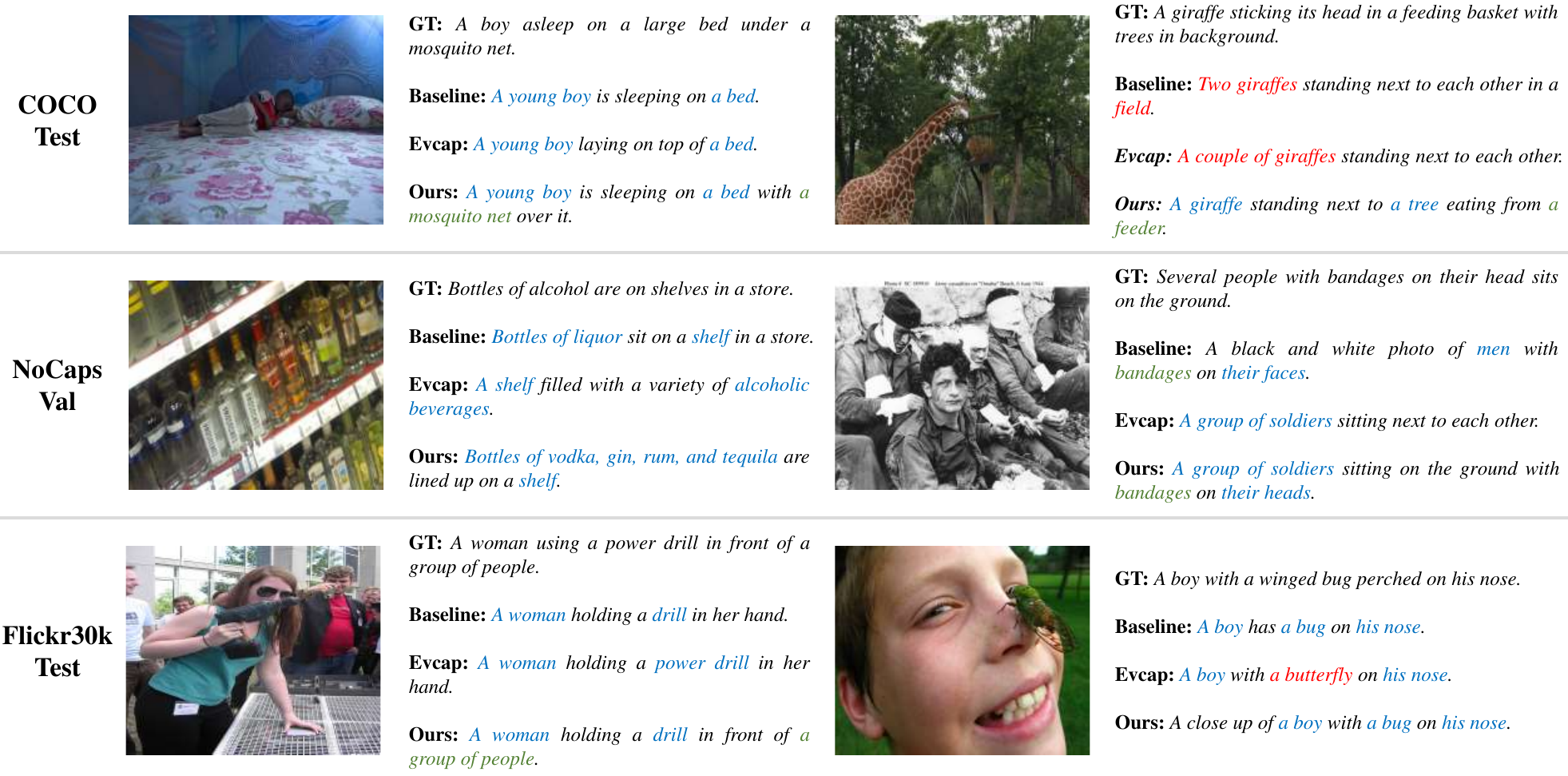}
	\caption{Examples of captions generated by our method, the baseline, and EVCap on regular datasets such as the COCO test set, NoCaps validation set, and Flickr30k test set. GT refers to the ground truth description. Incorrect objects in the caption are highlighted in \textcolor{red}{red}, correct objects in \textcolor[rgb]{0, 0.44, 0.75}{blue}, and partially ignored objects in \textcolor[rgb]{0.33, 0.51, 0.21}{green}. Our method generates accurate captions across different datasets.}
	\label{fig_4}
\end{figure*}

As shown in Fig.~\ref{fig_4}, to better analyze our model, we compare the results generated by the baseline, EVCap, and our model on the COCO test, NoCaps validation, and Flickr30k test. 

\textbf{Comparison on COCO test.} Our model described the mosquito net and feeder, which the baseline and EVCap did not describe, while the baseline and EVCap described the wrong number of giraffes. This indicates that our model has better alignment ability and can more accurately align visual content with textual information in LLMs.

\textbf{Comparison on the NoCaps validation.} All models perform well on NoCaps, but compared with the baseline and EVCap, our model describes more specific types of wine, indicating that our trigger-augmented module is effective. 

\textbf{Comparison on Flickr30k test.} There is little difference in performance among the models, but EVCap incorrectly describes a bug as a butterfly. The reason is that the additional retrieval library in EVCap generates incorrect information, which misleads the LLMs.

In conclusion, on the regular dataset, the results are not much different, and EVCap is affected by the retrieval database, occasionally generating incorrect entity objects. The baseline also has some entity errors without trigger-augmented module 1. Our model performs the best, accurately describing all the objects in the image and also providing detailed descriptions, which demonstrates the superiority of our model.

\subsubsection{Visualization on WHOOPS}
In order to better demonstrate the open-world recognition ability of our model, we present the experimental results on the unconventional dataset WHOOPS, as shown in Fig.~\ref{fig_5}. 

\textbf{Compared to the ground truth}, our model generates descriptions that are almost identical, demonstrating its superiority in describing open worlds. Meanwhile, the wolf and lamb lying together in the first image, the blue Pikachu in the second image, and the Statue of Liberty and the Sydney Opera House together in the fifth image are all unintuitive images that our model is still able to describe accurately. This indicates that our model accurately follows visual information for image description, rather than relying on the logical bias of large language models. 

\textbf{Compared to the baseline}, our description is more accurate. For example, in the description of the second image, Pikachu is replaced by the accurate blue Pikachu (the default Pikachu is yellow). In the third image, the Eiffel Tower replaces a generic tower. And in the fourth image, Einstein replaces a man. Finally, in the fifth image, the Sydney Opera House replaces a building. In particular, there are no aforementioned entities in the COCO dataset we trained on, which indicates that our model has zero-shot capability. This also demonstrates the effectiveness of our trigger-augmented module 1.

\section{Conclusion}
This paper presents TPCap, a retrieval-augmented image captioning framework that eliminates external retrieval banks by leveraging trigger-augmented generation and multi-modal purification. The trigger-augmented (TA) module activates LLMs' zero-shot capabilities while ensuring robust visual-textual alignment and mitigating bias. The multi-modal purification (MP) module refines generated information by filtering noise and enhancing entity relevance.
Evaluations on COCO, NoCaps, Flickr30k, and WHOOPS show that TPCap achieves competitive performance with only 0.82M trainable parameters, trained on a single NVIDIA RTX 4090 GPU. These results highlight the efficiency of leveraging LLMs' zero-shot capabilities for image captioning, offering a lightweight and scalable alternative to traditional retrieval-based methods.


\bibliographystyle{IEEEtran}
\bibliography{ref}

\newpage

\section{Biography Section}
If you have an EPS/PDF photo (graphicx package needed), extra braces are
 needed around the contents of the optional argument to biography to prevent
 the LaTeX parser from getting confused when it sees the complicated
 $\backslash${\tt{includegraphics}} command within an optional argument. (You can create
 your own custom macro containing the $\backslash${\tt{includegraphics}} command to make things
 simpler here.)

\vspace{11pt}

\bf{If you will not include a photo:}\vspace{-33pt}
\begin{IEEEbiographynophoto}{John Doe}
Use $\backslash${\tt{begin\{IEEEbiographynophoto\}}} and the author name as the argument followed by the biography text.
\end{IEEEbiographynophoto}

\vfill

\end{document}